\documentclass{article}
\usepackage{ijcai23}

\usepackage{times}
\usepackage{soul}
\usepackage{url}
\usepackage[hidelinks]{hyperref}
\usepackage[utf8]{inputenc}
\usepackage[small]{caption}
\usepackage{graphicx}
\usepackage{amsmath}
\usepackage{amsthm}
\usepackage{booktabs}
\usepackage[switch]{lineno}
\usepackage{booktabs}
\usepackage{amsfonts}
\usepackage[ruled]{algorithm2e}
\usepackage{multirow}
\usepackage{color}
\usepackage{lipsum}
\title{Incomplete Multi-view Clustering via Prototype-based Imputation}

\author{
Haobin Li$^{1,\dag}$
\and
Yunfan Li$^{1,\dag}$\and
Mouxing Yang$^1$\and
Peng Hu$^1$\and
Dezhong Peng$^1$\And
Xi Peng$^{1,*}$
\affiliations
$^1$College of Computer Science, Sichuan University 
\emails
\{haobinli.gm, yunfanli.gm, yangmouxing, penghu.ml\}@gmail.com,
pengdz@scu.edu.cn,
pengx.gm@gmail.com
}
\begin{document}

\maketitle
\newcommand\blfootnote[1]{%
\begingroup
\renewcommand\thefootnote{}\footnote{#1}%
\addtocounter{footnote}{-1}%
\endgroup
}
\begin{abstract}
In this paper, we study how to achieve two characteristics highly-expected by incomplete multi-view clustering (IMvC). Namely, i) instance commonality refers to that within-cluster instances should share a common pattern, and ii) view versatility refers to that cross-view samples should own view-specific patterns. To this end, we design a novel dual-stream model which employs a dual attention layer and a dual contrastive learning loss to learn view-specific prototypes and model the sample-prototype relationship. When the view is missed, our model performs data recovery using the prototypes in the missing view and the sample-prototype relationship inherited from the observed view.
Thanks to our dual-stream model, both cluster- and view-specific information could be captured, and thus the instance commonality and view versatility could be preserved to facilitate IMvC. Extensive experiments demonstrate the superiority of our method on six challenging benchmarks compared with 11 approaches. The code will be released.
\end{abstract}
\section{Introduction}
As a fundamental tool in multi-view data analysis, multi-view clustering (MvC) aims at partitioning instances into different clusters without the help of data annotations~\cite{tao2017ensemble,hu2019multi,huang2019multi,kang2020partition,yang2021deep}. Almost all of MvC works implicitly or explicitly take the data completeness assumption, \textit{i.e.}, all instances exist in all views. In practice, however, the assumption is always violated due to the complexity of data collection and transmission, leading to the incomplete problem in multi-view data. For example, when building medical history for patients, the multi-view healthcare data is susceptible to be incomplete due to disease concealment in the data collection or information loss during treatment transfer.
\blfootnote{$\dag$ Equal Contribution}
\blfootnote{$*$ Correspoonding Author}

\begin{figure}[t]
    \centering
    \includegraphics[width=0.49\textwidth]{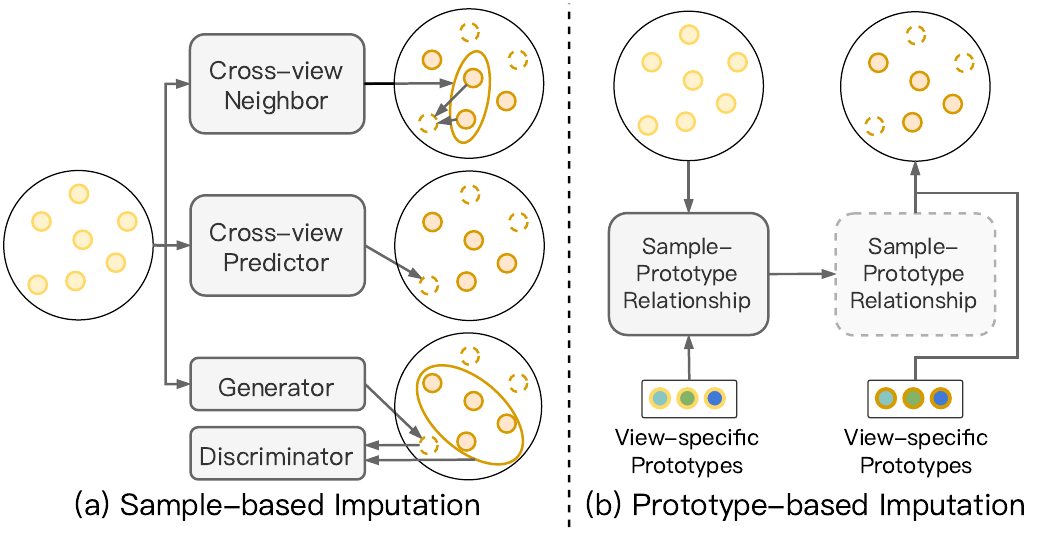}
    \caption{Our basic idea. (a) Three typical sample-based data recovery paradigms in existing IMvC studies, namely, i) neighborhood-based recovery, ii) cross-view prediction, and iii) adversarial generation. One limitation of the three paradigms is that two highly-expected characteristics in IMvC, \textit{i.e.}, instance commonality and view versatility, are not fully explored. (b) The proposed prototype-based imputation paradigm. In brief, the data is recovered using the sample-prototype relationship inherited from the observed view and the prototypes from the missing view. Such a paradigm could recover both cluster- and view-specific information, thus preserving instance commonality and view versatility.
    }
    \label{fig: idea}
\end{figure}

To achieve incomplete multi-view clustering (IMvC), a feasible solution is employing the observed cross-view samples to recover the missing counterparts and then performing clustering. As shown in Fig~\ref{fig: idea}(a), one of the most straightforward paradigms is using observed samples to find cross-view neighbors which are further used to recover missing samples. Such a paradigm implicitly assumes that the views could be mapped into a common space wherein the neighbors of the missing sample could be accurately identified by its cross-view counterpart. In practice, however, such an assumption is satisfied always at cost of the \textit{view versatility} since the view-specific information is often excluded to learn the common space. 
To compensate for view versatility, some studies propose capturing the view-specific information using a cross-view predictor~\cite{lin2021completer} or generator~\cite{wang2018partial}. Unfortunately, such a generative paradigm essentially learns an equivalent mapping for the whole dataset across views, which will lose the \textit{instance commonality}, \textit{i.e.}, within-cluster compactness and between-cluster scatterness. 

Different from the aforementioned sample-based imputation methods, we propose a prototype-based imputation paradigm as shown in Fig.~\ref{fig: idea}(b). Unlike existing methods that restore the missing sample through learning a common representation for cross-view samples, we propose performing data recovery using the prototypes from the missing view and the sample-prototype relationship from the observed view. Thanks to our paradigm, the instance commonality and the view versatility can be preserved because the prototypes capture the cluster- and view-specific information. Furthermore, our invariance assumption on the sample-prototype relationship is milder than that on the cross-view representation taken in these works. 

To implement the prototype-based imputation, ones have to overcome the following two technical challenges, \textit{i.e.}, i) incorporating prototypes and samples to enhance the instance commonality, and ii) learning view-specific prototypes to preserve view versatility. To this end, we propose an incomplete multi-view clustering method based on a novel dual-stream model consisting of a dual attention layer and a dual contrastive learning loss. To be specific, the dual attention layer aims to enhance the instance commonality by representing samples and prototypes with each other. More specifically, the sample representation is learned by aggregating the sample itself and the corresponding prototype, thus enhancing the commonality of with-cluster instances. In a dual manner, the prototype representation is learned through aggregating prototype itself and the current input samples, thus integrating the historical and current information. 
The dual contrastive learning loss is designed to preserve view versatility, which consists of the standard contrastive learning on samples and a new bounded contrastive loss on the prototypes. Thanks to the bounded contrastive loss, the prototypes will embrace the unique view-specific information, thus preserving the view versatility. The major contributions of this paper could be summarized as follows:
\begin{enumerate}
    \item From the standpoint of data recovery for IMvC, we proposed a novel imputation method which restores the missing samples using the prototypes and the sample-prototype relationship. 
    % takes a milder assumption compared with existing sample-based imputation methods, thus enjoying better data restoration. 
    % In brief, we assume the within-cluster sample-prototype relationship is invariant, while the methods assume that the cross-view representation is invariant. 
    Such a prototype-based imputation paradigm could preserve instance commonality and view versatility that are favorite to IMvC.
    % almost all existing works restore the missing data through mining invariant cross-views representation. 
    % In contrast, we exploit the invariant cross-views sample-prototype relationship. The  assumption taken in our method is milder than that 
    \item From the standpoint of unsupervised multi-view representation learning, we propose a novel dual-stream model which learns sample representation using prototypes and prototype representation using the input samples. Thanks to the dual-stream model, our method could learn better representation for boosting IMvC performance.  
    % which incorporates prototype information into samples to enhance commonality between within-cluster samples, as well as incorporates sample information into prototypes to 
    \item Extensive experiments on six benchmarks demonstrate the superiority of our method in both incomplete multi-view clustering and data recovery performance, compared with 11 baselines.
\end{enumerate}

\section{Related Work}
In this section, we briefly review two related topics, namely, incomplete multi-view clustering and attention-based model.

\subsection{Incomplete Multi-view Clustering}
IMvC is a long-standing task in the multi-view learning community, which has attracted numerous studies.
Based on the way to utilize the cross-view information, classic IMvC methods could be divided into three categories, including matrix factorization based~\cite{li2014partial,zhao2016incomplete,shao2015multiple,hu2019doubly}, kernel learning based~\cite{bach2002kernel,liu2020efficient}, and similarity relation based~\cite{wang2019spectral,liu2019multiple}. To handle more complex and large-scale data, several deep IMvC methods have been developed recently. Based on the paradigm of recovering the missing data, deep IMvC methods could be divided into three categories, including i) neighborhood-based methods~\cite{tang2022deep,yang2022robust}, which impute the missing data with the help of cross-view nearest neighbors, ii) predictor-based methods~\cite{lin2021completer,lin2022dual}, which learn a direct mapping from observed views to missing views for data recovery, and iii) GAN-based methods~\cite{wang2018partial,jiang2019dm2c,zhang2020deep}, which recover the missing data through adversarial generation.

Among the above works, deep IMvC methods are most similar to this study. However, this study is remarkably different from existing works in the following aspects. First, the existing works impute data based on the observed counterparts which might discard either instance commonality or view versatility. In  contrast, the proposed prototype-based imputation paradigm performs recovery using the prototypes in the missing view and the sample-prototype relationship in the observed view, thus taking the best of both worlds. Second, to the best of our knowledge, this could be the first attention-based model in the IMvC community, showing its great potential in unsupervised data recovery and IMvC. 
% In brief, we design a novel dual-stream model to simultaneously facilitate clustering-favorable feature extraction and missing data recovery.

\begin{figure*}[t]
    \centering
    \includegraphics[width=0.97\textwidth]{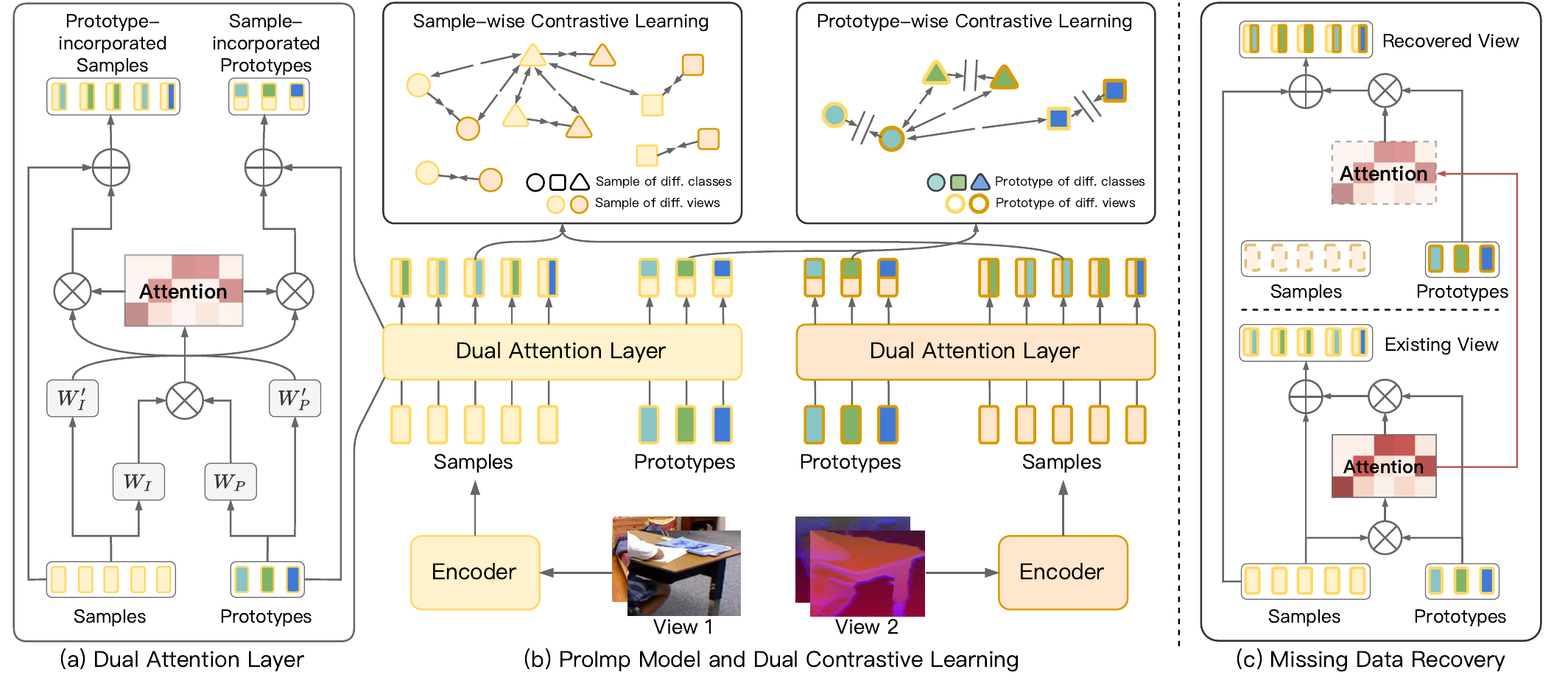}
    \caption{Overview of our ProImp method. (a) The dual attention layer. The attention is computed between samples and prototypes to incorporate each other. On the one hand, the sample representation aggregates the sample itself and the corresponding prototype, thus enhancing the commonality of with-cluster instances. On the other hand, the prototype representation aggregates the prototype itself and the current input samples, thus integrating the historical and current information. (b) The ProImp model and dual contrastive learning objective. To optimize the entire model as well as learnable view-specific prototypes, in addition to conducting standard contrastive learning on samples, we dually contrast prototypes with a new bounded contrastive loss to preserve view versatility. (c) The prototype-based missing data recovery. The missing samples are recovered with the attention inherited from the observed view and prototypes in the missing view, which enjoys both instance commonality and view versatility. Meanwhile, samples from the observed view are skip-connected to introduce instance consistency in the recovered data.}
    \label{fig: method}
\end{figure*} 

\subsection{Attention-Based Model}
The attention-based model learns better representation by focusing on regions with relevant information, which has achieved great success in various tasks such as image classification~\cite{yu2018rethinking}, person re-identification~\cite{hou2019interaction}, object detection~\cite{woo2018cbam}, neural machine translation~\cite{vaswani2017attention}, and sentence summarization~\cite{rush2015neural}. Recently, some works have explored the attention mechanism in multi-view learning. For example,~\cite{qu2017attention} promotes using attention to collaborate different views for multi-view representation learning. \cite{zhou2018aware} proposes a viewpoint-aware attention model for vehicle re-identification. \cite{luo2020attention} implements attention-enhanced matching confidence volume in multi-view stereo. \cite{yan2022multiview} introduces lateral connections by cross-view attention, and fuses multi-view information for video recognition.

The major differences between this work and previous attention-based models lie in two aspects. First,  different from most existing works that focus on single-stream and instance-wise attention, the proposed dual-stream model employs a novel dual attention layer to incorporate samples and learnable prototypes with each other. Second, unlike most existing works that solely use attention for general multi-view feature fusion, the proposed dual attention layer is IMvC-oriented, which simultaneously facilitates clustering-favorable feature extraction and data recovery.

\section{Method}
In this section, we propose a dual-stream model dubbed ProImp to achieve incomplete multi-view clustering. As illustrated in Fig.~\ref{fig: method}, ProImp is composed of a dual attention layer to model the relationship between samples and prototypes, as well as a dual contrastive learning loss to learn attention and view-specific prototypes. For data recovery, ProImp adopts the prototype-based imputation paradigm to preserve instance commonality and view versatility. In the following, we first introduce our dual attention layer in Sec.~\ref{sec: att}, then elaborate on the dual contrastive learning loss in Sec.~\ref{sec: optimize}, and finally present the prototype-based imputation paradigm in Sec.~\ref{sec: impute}.

\subsection{Dual Attention Layer}
\label{sec: att}
Without loss of generality, we take bi-view data as an example for clarity. Let $\mathrm{\mathbf{X}}=\left\{\mathrm{\mathbf{X}}^{1, 2}, \mathrm{\mathbf{X}}^1, \mathrm{\mathbf{X}}^2\right\}$ be an incomplete multi-view dataset, where $\mathrm{\mathbf{X}}^1, \mathrm{\mathbf{X}}^2, \mathrm{\mathbf{X}}^{1, 2}$ refer to three subsets of instances that have data observed in the first, the second, and both views. We denote the set of $N$ complete instances as $\mathrm{\mathbf{X}}^{1, 2} = \{X^1, X^2\}$, where $X^v = \{x^v_1, x^v_2, \dots, x^v_N\}$ denotes the samples in the $v$-th view.

As illustrated in Fig.~\ref{fig: method}, the dual attention is computed between samples $X^v$ and a set of learnable prototypes $C^v = \{c^v_1, c^v_2, \dots, c^v_K\}$, where $K$ corresponds to the target cluster number. Mathematically, the attention $A^v$ is computed through
\begin{equation}
    A^v = \operatorname{Softmax}\left({(W^v_I X^v)^T W^v_P C^v} / {\sqrt{d}}\right)
\label{eq: att}
\end{equation}
where $W^v_I$ and $W^v_P$ are two linear layers for samples and prototypes in $v-$th view respectively, and $d$ is the dimension of features.

The attention $A^v$ is then used to incorporate samples and prototypes in a dual manner. For sample representation, the corresponding prototype is aggregated to each sample, namely,
\begin{equation}
    Z^v=X^v+ A^v {W^\prime}^v_P C^v,
\end{equation}
where $Z^v = \{z^v_1, z^v_2, \dots, z^v_N\}$ is the new representation of samples, and ${W^\prime}^v_P$ is another linear layer for prototypes. Such behavior intrinsically pulls each sample to its corresponding cluster center, thus enhancing the instance commonality favored in clustering.

Likewise, for prototype representation, features of current samples would be aggregated into prototypes, namely,
\begin{equation}
    U^v=C^v+ A^v {W^\prime}^v_I X^v
\end{equation}
where $U^v = \{u^v_1, u^v_2, \dots, u^v_K\}$ is the new representation of prototypes, and ${W^\prime}^v_I$ is another linear layer for samples. Such behavior enables prototypes to integrate the historical and current cluster information.

Notably, an encoder network is adopted to extract the features of samples in each view before feeding them to the dual attention layer. Here we omit it in mathematical notations for simplicity.

\subsection{Dual Contrastive Learning}
\label{sec: optimize}
As discussed above, the dual attention layer outputs prototype-incorporated samples and sample-incorporated prototypes in each view. To optimize the entire model and learnable view-specific prototypes, we conduct dual contrastive learning on samples and prototypes, respectively.

\subsubsection{Sample-wise contrastive learning}
To mine instance consistency between cross-view samples, we adopt the following contrastive loss that maximizes the similarities between cross-view samples of the same instance, while minimizing those between samples of different instances, namely,
\begin{align}
    \mathcal{L}_{S}=\frac{1}{2N}\sum_{i=1}^{N}\left(\mathcal{L}_{i}^{1,2}+\mathcal{L}_{i}^{2,1}\right),~~~~~~~~~~~~~~~\\
    \mathcal{L}_{i}^{1,2}=
        -\log \frac{e^{s\left(z_i^{1}, z_i^{2}\right) / \tau_{I}}}{\sum_{\substack{j=1}}^N \left[e^{s\left(z_i^{1}, z_j^{1}\right)/ \tau_{I}}+ e^{s\left(z_i^{1}, z_j^{2}\right) / \tau_{I}}\right]},
\label{eq: isl}
\end{align}
where $s\left(\cdot, \cdot\right)$ denotes the cosine similarity, $\tau_{I}=0.5$ is the temperature parameter, and $\mathcal{L}_{i}^{2,1}$ is defined similarly as $\mathcal{L}_{i}^{1,2}$.

\subsubsection{Prototype-wise contrastive learning}
As discussed, our prototype-based imputation paradigm requires prototypes to capture view versatility. In other words, prototypes from different views should not collapse into an identical representation. To this end, instead of simply maximizing the similarities between cross-view prototypes of the same cluster, we propose to optimize their similarities to a bound. Meanwhile, to achieve a more distinct clustering, we minimize the similarities between prototypes of different clusters, which leads to the following bounded contrastive loss,
\begin{equation}
\begin{aligned}
    \mathcal{L}_{P} =& \frac{2}{K} \sum_{i=1}^{K} \frac{\left|s\left(u_i^{1}, u_i^{2}\right)-\alpha \right|}{\tau_{P}} + \frac{1}{K} \sum_{\substack{v_1=1 \\ v_2=1}}^2 \sum_{i=1}^K \\
    & \log \left[ e^{\left|s\left(u_i^{v_1}, u_i^{v_2}\right)-\alpha \right| / \tau_{P}} + \sum_{\substack{j=1, j\neq i}}^K e ^{s\left(u_i^{v1}, u_j^{v2}\right) / \tau_{P}} \right],
\end{aligned}
\label{eq: psl}
\end{equation}
where $\alpha$ denotes the similarity bound and $\tau_{P}=2.0$ is the temperature parameter.

\subsubsection{Attention Regularization}
Recall that the dual attention $A^v$ defined in Eq.~\ref{eq: att} is an $N\times K$ matrix, where $A^v_{ij}$ intrinsically corresponds to the probability of the $i$-th sample belonging to the $j$-th cluster. To achieve more distinct clustering, we expect each sample to be confidently assigned to a certain cluster. Meanwhile, we should prevent the trivial solution where most samples are assigned to the same cluster. For these purposes, we propose the following attention regularization term, namely,
\begin{equation}
\label{eq: reg}
    \mathcal{L}_{R} = \sum_{v=1}^{2} \sum_{j=1}^{K}\left[ A_{\cdot j}^{v}\log A_{\cdot j}^{v} - \beta \sum_{i=1}^{N} A_{ij}^{v}\log A_{ij}^{v} \right],
\end{equation}
where $A_{\cdot j}^{v} = \sum_{i=1}^{N} A_{ij}^{v}$ and $\beta$ is the weight parameter to balance the sharpness and uniformity of attention.

Combining the dual contrastive learning loss and the attention regularization term, we arrive at the overall objective function of the proposed ProImp, namely,
\begin{equation}
\mathcal{L} = \mathcal{L}_{S} + \mathcal{L}_{P} + \mathcal{L}_{R}.
\label{eq: overall}
\end{equation}

\subsection{Prototype-based Imputation}
\label{sec: impute}
To recover the missing samples in the incomplete data $\{\mathrm{\mathbf{X}}^1, \mathrm{\mathbf{X}}^2\}$, we design the following prototype-based imputation strategy as illustrated in Fig.~\ref{fig: method}(c). Specifically, given data $\mathrm{\mathbf{X}}^1$ observed in view 1, the missing data in view 2 is recovered with attention $A^1$ and prototypes $C^2$ through
\begin{equation}
\label{eq: impute}
    \hat{Z}^2 = X^1 + A^1 {W^\prime}^2_P C^2,
\end{equation}
where $A^1$ is the dual attention computed in view 1 according to Eq.~\ref{eq: att}, and $\hat{Z}^2$ is the recovered data in view 2. The idea behind such an attention inheritance is that the instance semantics are expected to be consistent across different views. By incorporating cluster- and view-specific information from prototypes, both instance commonality and view versatility could be preserved in the recovered data. In addition, samples from the observed view are skip-connected to the recovered data to introduce instance consistency.

Likewise, the missing data $\hat{Z}^1$ in view 1 is similarly recovered given data $\mathrm{\mathbf{X}}^2$ observed in view 2. Let $\textbf{Z}^1$ and $\textbf{Z}^2$ denote the observed views in the incomplete data, the representation $\mathrm{\mathbf{Z}} = \{ \{Z^1, Z^2 \}, \{ \textbf{Z}^1, \hat{Z}^2 \}, \{ \hat{Z}^1, \textbf{Z}^2 \} \}$ of both the observed and recovered data is concatenated and fed into the k-means algorithm to achieve clustering. Notably, though the attention itself intrinsically corresponds to the cluster assignment, it only utilizes data from a single view. Therefore, a simple concatenation operation is applied to gather multi-view information.

\section{Experiment}
In this section, we evaluate the proposed ProImp method on six widely-used multi-view datasets compared with 11 baselines. First, we present the experimental setting and implementation details in Sec.~\ref{sec: setting}. Then, we compare our ProImp with state-of-the-art methods in Sec.~\ref{sec: sota}. After that, we conduct the parameter analyses and ablation studies in Sec.~\ref{sec: ablation}. Finally, we present visualization results in Sec.~\ref{sec: visulization}.

\begin{table*}[t]
\caption{The clustering performance on four multi-view benchmarks. The best and second best results are denoted in \textbf{bold} and \underline{underline}.}
\label{tab: main}
\resizebox{\textwidth}{!}{
\begin{tabular}{c|l|ccc|ccc|ccc|ccc}
\hline
\multirow{2}{*}{Setting} & \multicolumn{1}{c|}{\multirow{2}{*}{Method}} & \multicolumn{3}{c|}{Scene-15} & \multicolumn{3}{c|}{Reuters} & \multicolumn{3}{c|}{NoisyMNIST} & \multicolumn{3}{c}{CUB} \\ & \multicolumn{1}{c|}{} & ACC & NMI & ARI & ACC & NMI & ARI & ACC & NMI & ARI & ACC & NMI & ARI \\ \hline
\multirow{13}{*}{Incomplete} & DCCA(ICML'13) & 28.78 & 28.35 & 13.24 & 45.84 & 26.08 & 18.00 & 63.75 & 61.72 & 41.17 & 44.20 & 43.30 & 26.65 \\
& DCCAE(ICML'15) & 29.01 & 29.13 & 12.86 & 47.04 & 28.00 & 14.48 & 65.42 & 62.87 & 38.32 & 42.33 & 40.87 & 25.46 \\
& BMVC(TPAMI'18) & 32.45 & 30.87 & 11.56 & 32.10 & 6.98 & 2.89 & 30.71 & 19.16 & 10.60 & 29.79 & 20.28 & 6.35 \\
& AE$^{2}$-Nets(CVPR'19) & 22.44 & 23.43 & 9.56 & 29.08 & 7.55 & 4.84 & 29.88 & 23.78 & 11.81 & 35.87 & 32.00 & 15.90 \\
& PMVC(AAAI'14) & 25.47 & 25.37 & 11.31 & 29.32 & 7.42 & 4.42 & 33.13 & 25.49 & 14.62 & 57.73 & 54.37 & 38.29 \\
& UEAF(AAAI'19) & 28.95 & 26.92 & 8.37 & 33.32 & 20.06 & 12.19 & 37.45 & 34.42 & 25.71 & 45.80 & 45.25 & 26.88 \\
& DAIMC(IJCAI'18) & 27.00 & 23.47 & 10.62 & 40.94 & 18.66 & 15.04 & 33.81 & 26.42 & 15.96 &   62.70 & 58.48 & 47.72 \\
& EERIMVC(TPAMI'20) & 31.50 & 31.11 & 14.82 & 29.77 & 12.01 & 4.21 & 55.62 & 45.92 & 36.76 &  \underline{68.73} & 63.90 & \underline{53.77} \\
& COMPLETER(CVPR'21) & 39.50 & \underline{42.35} & \underline{23.51} & 34.61 & 17.53 & 2.93 & 80.01 & 75.23 & 70.66 & 53.66 & \underline{65.45} & 47.26 \\
& SURE(TPAMI'22) & \underline{39.60} & 41.58 & 23.49 & \underline{47.18} & \underline{30.89} & \underline{23.32} & \underline{92.34} & \underline{84.99} & \underline{84.31} & 58.33 & 50.37 &37.44 \\
& DSIMVC(ICML'22) &30.56 &35.47 &17.24 &39.87 &19.61 & 17.13 &57.47 & 55.12 &44.08 &   54.57 &   51.35 & 35.04 \\
& \textbf{ProImp(Ours)} & \textbf{41.58} & \textbf{42.86} & \textbf{25.31} & \textbf{51.89} & \textbf{35.54} & \textbf{28.53} & \textbf{94.86} & \textbf{87.43} & \textbf{89.08} & \textbf{73.30} & \textbf{66.38} & \textbf{54.84} \\ \hline
\multirow{13}{*}{Complete} & DCCA(ICML'13) & 36.61 & 39.20 & 21.03 & 47.95 & 26.57 & 12.71 & 89.64 & 88.33 & 83.95 & 55.60 & 56.11 & 43.18 \\
& DCCAE(ICML'15) & 34.58 & 39.01 & 19.65 & 41.98 & 20.30 & 8.51 & 78.00 & 81.24 & 68.15 & 55.30 & 58.70 & 45.05 \\
& BMVC(TPAMI'18) & 40.50 & 41.20 & 24.11 & 42.39 & 21.86 & 15.14 & 88.31 & 77.01 & 76.58 & 66.21 & 61.70 & 48.69 \\
& AE$^{2}$-Nets(CVPR'19) & 37.17 & 40.47 & 22.24 & 42.39 & 19.76 & 14.87 & 52.83 & 51.24 & 39.52 & 48.80 & 46.71 & 30.49 \\
& PMVC(AAAI'14) & 30.83 & 31.05 & 14.98 & 32.50 & 11.11 & 7.48 & 41.09 & 36.36 & 24.47 & 64.53 & 70.34 & 53.11 \\
& UEAF(AAAI'19) & 34.37 & 36.69 & 18.52 & 40.19 & 24.34 & 15.94 & 66.22 & 64.34 & 54.83 &   63.33 & 56.91 & 44.48 \\
& DAIMC(IJCAI'18) & 32.09 & 33.55 & 17.42 & 40.78 & 21.15 & 15.98 & 38.40 & 34.66 & 22.98 &  71.57 & 70.69 & 57.89 \\
& EERIMVC(TPAMI'20) & 39.60 & 38.99 & 22.06 & 33.21 & 14.28 & 3.90 & 65.66 & 57.60 & 51.34 & \underline{74.00} & \underline{73.05} & \underline{62.41}\\
& COMPLETER(CVPR'21)  & \underline{41.07} & \underline{44.68} & 24.78 & 36.20 & 18.87 & 4.75 & 89.08 & 88.86 & 85.47 & 63.57 & 70.18 & 52.92 \\
& SURE(TPAMI'22) & 40.95 & 43.19 & \underline{25.01} & \underline{49.06} & \underline{29.91} & \underline{23.56} & \underline{98.36} & \underline{95.38} & \underline{96.43} & 58.00 & 59.32 & 45.16 \\
& DSIMVC(ICML'22) & 31.66 & 35.61 & 17.21 & 43.20 & 23.29 & 19.02 & 60.98 & 58.09 & 46.74 & 59.67 & 57.12 & 41.26 \\
& \textbf{ProImp(Ours)} & \textbf{43.61} & \textbf{45.02} & \textbf{26.84} & \textbf{56.54} & \textbf{39.35} & \textbf{32.77} & \textbf{99.17} & \textbf{97.48} & \textbf{98.18} & \textbf{80.63} & \textbf{75.48} & \textbf{66.04} \\ \hline
\end{tabular}}
\end{table*}

\begin{figure*}[t]
    \centering
    \includegraphics[width=1.00\textwidth]{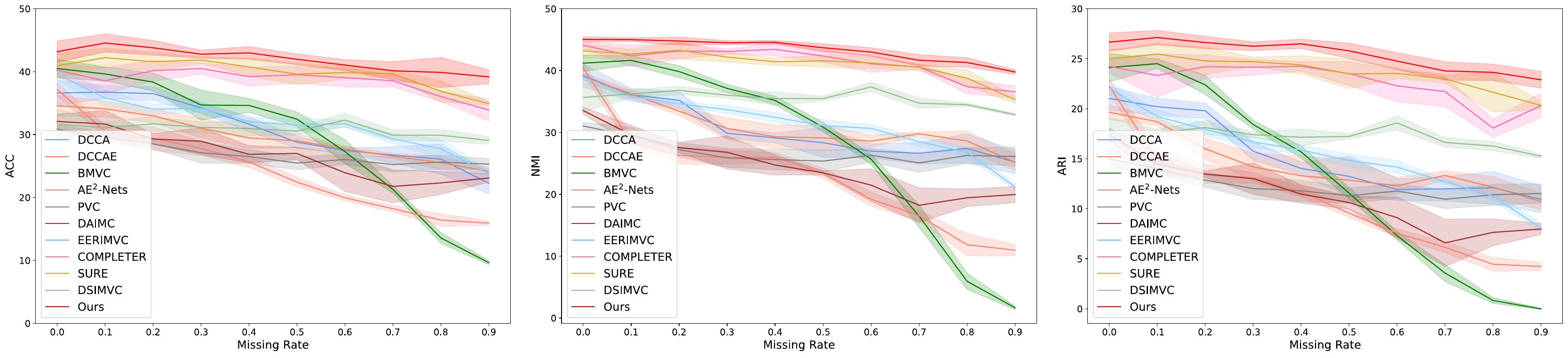}
    \caption{Clustering performance on Scene-15 under different missing rates. The colored regions denote the standard variances in five random experiments.}
    \label{fig: missing rate}
\end{figure*}

\begin{figure*}[t]
    \centering
    \includegraphics[width=0.99\textwidth]{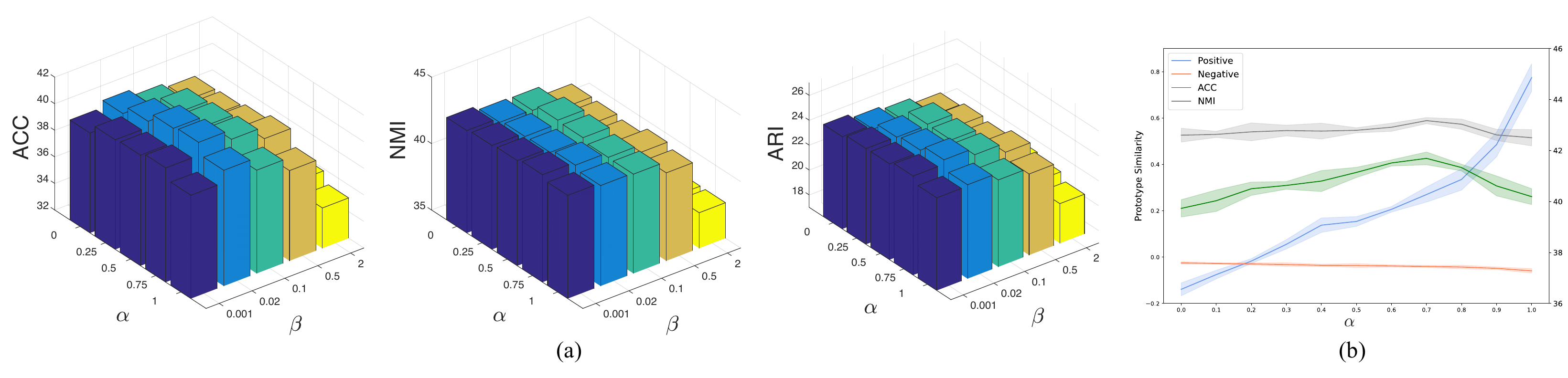}
    \caption{Parameter analyses on Scene-15. (a) The clustering performance of ProImp under different choices of the similarity bound $\alpha$ and the balance weight $\beta$. (b) The cross-view prototype similarity and clustering performance under different choices of $\alpha$.}
    \label{fig: param}
\end{figure*}

\subsection{Experimental Settings}
\label{sec: setting}
Six multi-view datasets are used in our experiments, including Scene15~\cite{fei2005bayesian}, Reuters~\cite{amini2009learning}, NoisyMNIST~\cite{wang2015deep}, CUB~\cite{wah2011caltech}, Deep Animal~\cite{zhang2020deep}, and MNIST-USPS~\cite{peng2019comic}. Among them, the experiments of Animal and MNIST-USPS are shown in the Supplementary Material. A detailed description of these datasets is provided in the Supplementary Material. We randomly remove one view for $m$ instances to simulate incomplete multi-view data with a missing rate of $m/n$, where $n$ corresponds to the total number of instances.

The proposed ProImp is implemented in PyTorch 1.11.0 and all the experiments are conducted on an NVIDIA 3090 GPU on Ubuntu 20.04 OS. The model is trained for 150 epochs using the Adam optimizer with an initial learning rate of 1e-3, with a batch size of 1,024 on all datasets.
The similarity bound $\alpha$ in Eq.~\ref{eq: psl} and the weight parameter in Eq.~\ref{eq: reg} are set to 0.75 and 0.02, respectively.
In practice, we first warm up the model with the sample-wise contrastive loss in Eq.~\ref{eq: isl} and the regularization term in Eq.~\ref{eq: reg} for 50 epochs. After that, we align the prototypes in different views with the Hungarian algorithm and train the model with the overall loss in Eq.~\ref{eq: overall} till the end.

Three widely-used metrics including Accuracy (ACC), Normalized Mutual Information (NMI), and Adjusted Rand Index (ARI) are used for evaluation. A higher value of these metrics indicates a better clustering performance.

\subsection{Comparisons with State of the Arts}
\label{sec: sota}
We compare ProImp with 11 state-of-the-art multi-view clustering baselines, including DCCA~\cite{andrew2013deep}, DCCAE~\cite{wang2015deep}, BMVC~\cite{zhang2018binary}, AE$^2$-Nets~\cite{zhang2019ae2}, PMVC~\cite{li2014partial}, UEAF~\cite{wen2019unified}, DAIMC~\cite{hu2019doubly}, EERIMVC~\cite{liu2020efficient}, COMPLETER~\cite{lin2021completer}, SURE~\cite{yang2022robust}, and DSIMVC~\cite{tang2022deep}. The implementation details for these baselines are provided in the Supplementary Material.

We first evaluate ProImp and baselines under the Incomplete (with the missing rate of 50\%) and Complete (with the missing rate of 0\%) scenarios. Table~\ref{tab: main} and Supplementary Table~1 show the average clustering performance under five random experiments. As can be seen, our ProImp significantly outperforms the state-of-the-art methods on all datasets. In particular, ProImp achieves a relatively 22\% (28.53\%~v.s.~23.32\%) and 39\% (32.77\%~v.s.~23.56\%) ARI improvement under the Incomplete and Complete scenarios on the Reuters dataset, compared with the second best method SURE. The superior performance demonstrates the effectiveness of the proposed dual-stream model, including the dual attention layer and dual contrastive learning objective.

We further explore the robustness of ProImp by increasing the missing rate from 0\% to 90\% with a gap of 10\% on the Scene-15 dataset. Considering that the number of complete instances would be greatly reduced under large missing rates, we adjust the batch size to 128 and set the learning rate as 3e-4 in this experiment. As shown in Fig.~\ref{fig: missing rate}, our ProImp substantially outperforms baselines under all missing rates. In addition, the performance of ProImp drops less as the miss rate increases. For example, in terms of ACC, ProImp outperforms SURE by 2.23\% (43.18\%~v.s.~40.95\%) under the complete scenario, and the performance gap increases to 4.26\% (39.17\%~v.s.~34.91\%) under 90\% missing rate. Such a result demonstrates the superiority of our attention imputation strategy for data recovery, as it could preserve both the view versatility and instance commonality information.
\begin{figure*}[t]
    \centering
    \includegraphics[width=0.97\textwidth]{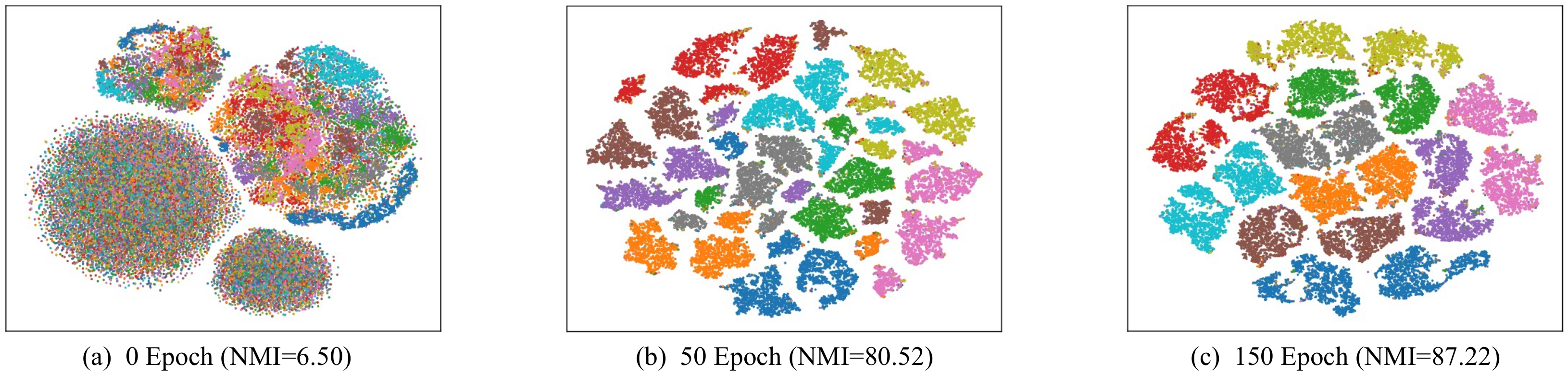}
    \caption{t-SNE visualization on the NoisyMNIST dataset across the training process.}
    \label{fig: epoch}
\end{figure*} 

\begin{figure}[t]
    \centering
    \includegraphics[width=0.45\textwidth]{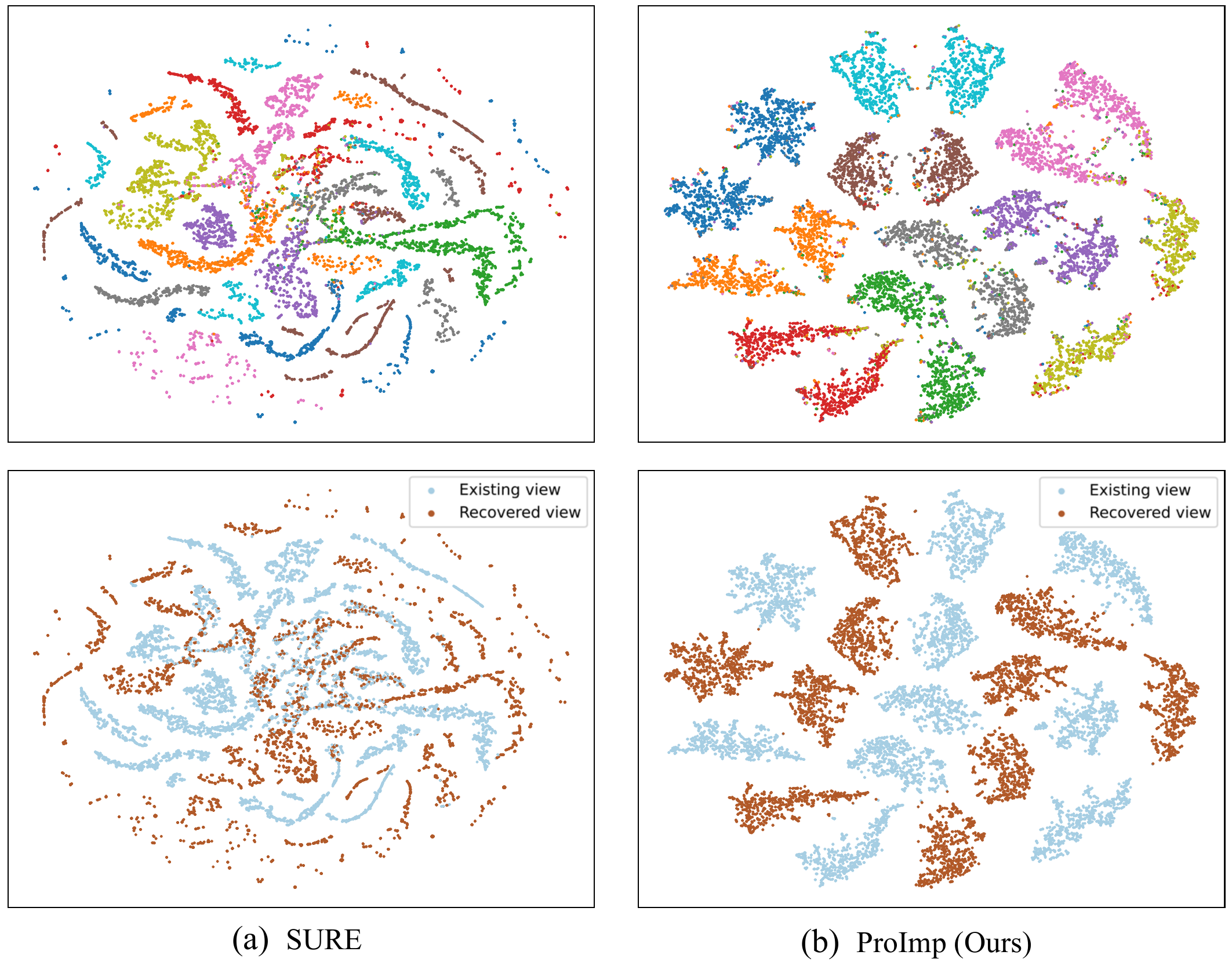}
    \caption{t-SNE visualization of observed and recovered data on NoisyMNIST, compared with the best competitor SURE. Data are colored by classes and views in the first and second rows, respectively.}
    \label{fig: recover}
\end{figure}
\subsection{Parameter Analyses and Ablation Studies}
\label{sec: ablation}

In this section, to better understand the effectiveness of each component in the proposed ProImp, we conduct a series of parameter analyses and ablation studies. In brief, we first investigate the influence of the similarity bound $\alpha$ in the prototype stream loss and the balance weight $\beta$ in the attention regularization term. Then we perform ablation studies on each loss term. Finally, we test variant data recovery strategies.

\subsubsection{Influence of hyper-parameters $\alpha$ and $\beta$}

There are two hyper-parameters in the proposed ProImp, namely, the similarity bound $\alpha$ in the prototype stream loss and the balance weight $\beta$ in the attention regularization term. To investigate how they influence the performance of ProImp, we change $\alpha$ in the range of $\{0,0.25,0.5,0.75,1\}$ and $\beta$ in the range of $\{0.001,0.02,0.1,0.5,2\}$. As shown in Fig.~\ref{fig: param}a, ProImp achieves the best performance when $\alpha=0.75$. According to the cross-view prototype similarity shown in Fig.~\ref{fig: param}b, positive prototype pairs get closer as $\alpha$ increases. An over-small cross-view prototype similarity would harm view consistency, and an over-large value would sacrifice the view versatility, both leading to inferior performance. As for the other parameter $\beta$, we find that ProImp achieves promising results under a reasonable range (\textit{i.e.}, from 0.001 to 0.1). However, when the balance weight is too large, the attention between each instance and prototype tends to be equal. Such a collapsed attention would cause a significant performance drop.

\subsubsection{Effectiveness of each loss term}

To explore the effectiveness of the proposed sample-wise contrastive loss, prototype-wise contrastive loss, and attention regularization, we conduct the ablation experiments on the three losses in Eq.~\ref{eq: overall}. According to the results shown in Table~\ref{tab: loss}, the regularization term $\mathcal{L}_{R}$ itself is not sufficient to learn appropriate attention. Both $\mathcal{L}_{S}$ and $\mathcal{L}_{P}$ could guide attention optimization, leading to better clustering performance. The best performance is achieved when all three losses are adopted, as both the instance commonality and view versatility are achieved.

\begin{table}[!h]
\caption{Ablation study of three losses on Scene-15, where "$\checkmark$" denotes the loss is adopted.}
\label{tab: loss}
\centering
\begin{tabular}{ccc|ccc}
\hline
$\mathcal{L}_{S}$ & $\mathcal{L}_{P}$ & $\mathcal{L}_{R}$ & ACC & NMI & ARI \\
\hline
& & $\checkmark$ & 25.12 & 23.91 & 10.72 \\
$\checkmark$ & & $\checkmark$ & 39.37 & 42.21 & 23.86 \\
& $\checkmark$ & $\checkmark$ & 27.73 & 25.29 & 12.20 \\
$\checkmark$ & $\checkmark$ & $\checkmark$ & \textbf{41.58} & \textbf{42.86} & \textbf{25.31} \\
\hline
\end{tabular}
\end{table}

\subsubsection{Variants of data recovery strategy}

Recall that to preserve instance commonality and view versatility, we recover the missing view by the sample-prototype attention inherited from the observed view and prototypes from the missing view, namely, $\hat{Z}^2 = X^1 + A^1 {W^\prime}^2_P C^2$ via Eq.~\ref{eq: impute}. Here, to prove the superiority of our paradigm, we further investigate three other variants of recovery strategies on the Scene-15 dataset. Specifically, 
\begin{itemize}
    \item Prototypes from observed views: recovering by using the prototypes and sample-prototype attention from the observed view, \textit{i.e.}, $\hat{Z}^2 = X^1 + A^1 {W^\prime}^1_P C^1$;
    \item Prototypes from missing views only: recovering by only using prototypes from the missing view, \textit{i.e.}, $\hat{Z}^2 = 2 A^1 {W^\prime}^2_P C^2$;
    \item Samples from observed views only: recovering by only using the observed cross-view counterparts, \textit{i.e.}, $\hat{Z}^2 = 2 X^1$;
    % \item Copy of observed view: directly copy the representation learned from the observed view, namely, $\hat{Z}^2 = X^1 + A^1 {W^\prime}^1_P C^1$;
    % \item Sample only: use only the view consistency information from the instance representation, namely, $\hat{Z}^2 = 2 X^1$;
    % \item Prototype only: use only the view complementarity information from the attention and prototype representation, namely, $\hat{Z}^2 = 2 A^1 {W^\prime}^2_P C^2$.
\end{itemize}
Notably, as both sample and prototype features are L2 normalized, we scale the features of the last two variants to keep the length consistent. From the results in Table~\ref{tab: recover}, one could have the following conclusions. First, replacing the missing view prototypes with those in the observed view would lose the view versatility, thus remarkably degrading the performance. Second, solely using prototypes or cross-view counterparts suffers from losing either cross-view consistency or instance commonality, resulting in sub-optimal results. In comparison, our default paradigm takes the best of both worlds, leading to the best performance.

\begin{table}[!h]
\caption{Ablation study on different data recovery strategies on Scene-15. ``P." denotes prototypes and ``S." denotes samples.}
\label{tab: recover}
\centering
\begin{tabular}{lccc}
\hline
Strategy & ACC & NMI & ARI \\ \hline
P. from observed views & 32.23 & 34.44 & 17.73   \\
% Direct Swap & 40.86 & 41.78 & 23.98 \\
P. from missing views only & 36.32 & 39.06 & 21.14 \\
S. from observed views only & 40.17 & 41.43 & 23.66 \\
\hline
Default & \textbf{41.58} & \textbf{42.86} & \textbf{25.31} \\
\hline
\end{tabular}
\end{table}

\subsection{Visualizations}
\label{sec: visulization}

In this section, we present two visualization results on the NoisyMNIST dataset with a missing rate of $0.5$ to provide an intuitive understanding of the training process and data recovery performance of ProImp.

\subsubsection{Features learned by ProImp across the training process}
We conduct t-SNE visualization on features learned by ProImp at three different training epochs in Fig.~\ref{fig: epoch}. As can be seen, data forms four clusters at the initialization, which corresponds to the observed and recovered data from two views. After 50 epochs, data tends to form semantic clusters. However, as the prototypes are not yet matched across views, the recovered data is not semantically aligned with the observed data. At the end of the training, the gap between observed and recovered within-cluster samples is significantly narrowed, indicating a good instance commonality. Meanwhile, samples from different views are still not collapsed together, indicating that the view versatility is well preserved.

\subsubsection{Data recovery performance}
As discussed, a major advantage of our prototype-based imputation strategy is that it could preserve both instance commonality and view versatility in the recovered data. To prove its superiority, we visualize the observed and recovered data learned by our ProImp and the best competitor SURE in Fig.~\ref{fig: recover}. From the results, one could see that i) data recovered by our ProImp forms more compact clusters, thanks to the dual-stream model which enhances the commonality between within-cluster instances, and ii) data recovered by our ProImp shows a more distinct pattern with observed data, which demonstrates that the view versatility is preserved from the view-specific prototypes.
\section{Conclusion}
To implement the proposed prototype-based imputation paradigm, we proposed a dual-stream model by designing a dual attention layer and a dual contrastive learning loss. Thanks to the proposed model, the instance commonality and view versatility could be preserved into representation, thus boosting the IMvC performance. Extensive experiment results demonstrate the superiority of our model in both clustering and data recovery performance. 
In the future, we plan to extend ProImp to handle the datasets that consist of three and more views.

%% The file named.bst is a bibliography style file for BibTeX 0.99c
\bibliographystyle{named}
\bibliography{ijcai23}

\end{document}